\documentclass[11pt,a4paper]{article}
\usepackage[hyperref]{emnlp2020}
\usepackage{times}
\usepackage{latexsym}

\usepackage{microtype}
\usepackage{microtype}
\usepackage{graphicx}
\usepackage{amsfonts,amssymb}
\usepackage{subfigure}
\usepackage{newunicodechar}
\usepackage{amsmath}
\usepackage{makecell}
\usepackage{multirow}
\usepackage{arydshln}
\usepackage{color,xcolor}

\aclfinalcopy 

\title{Attention Is All You Need for Chinese Word Segmentation}

\author{Sufeng Duan$^{1,2,3}$, Hai Zhao$^{1,2,3}$\thanks{$^{*}$Corresponding author. This paper was partially supported by National Key Research and Development Program of China (No. 2017YFB0304100), Key Projects of National Natural Science Foundation of China (U1836222 and 61733011), Huawei-SJTU long term AI project, Cutting-edge Machine reading comprehension and language model.}  \\
	$^{1}$Department of Computer Science and Engineering, Shanghai Jiao Tong University \\
	$^{2}$Key Laboratory of Shanghai Education Commission for Intelligent Interaction \\ and Cognitive Engineering, Shanghai Jiao Tong University, Shanghai, China\\
	$^{3}$MoE Key Lab of Artificial Intelligence, AI Institute, Shanghai Jiao Tong University \\
{\tt 1140339019dsf@sjtu.edu.cn, zhaohai@cs.sjtu.edu.cn} \\}

\date{}

\begin{document}
\maketitle
\begin{abstract}
Taking greedy decoding algorithm as it should be, this work focuses on further strengthening the model itself for Chinese word segmentation (CWS), which results in an even more fast and more accurate CWS model. Our model consists of an attention only stacked encoder and a light enough decoder for the greedy segmentation plus two highway connections for smoother training, in which the encoder is composed of a newly proposed Transformer variant, Gaussian-masked Directional (GD) Transformer, and a biaffine attention scorer. With the effective encoder design, our model only needs to take unigram features for scoring. Our model is evaluated on SIGHAN Bakeoff benchmark datasets. The experimental results show that with the highest segmentation speed, the proposed model achieves new state-of-the-art or comparable performance against strong baselines in terms of strict closed test setting.
\end{abstract}

\section{Introduction}
Chinese word segmentation (CWS) is the task of delimiting word boundaries in a sentence, as a basic and essential task for Chinese and many other East Asian languages which are written without explicit word delimiters, and thus different from alphabetical languages like English.  

\begin{table*}[!htb]

\centering
\small
\scalebox{0.9}{
\begin{tabular}{|l|l|l|l|}
\hline
&\multicolumn{1}{c|}{\textbf{ Traditional Models}} & \multicolumn{1}{c|}{\textbf{Neural Models}}&  \makecell[l]{ \textbf{Decoding} \\ \textbf{Algorithm}}\\ 
\hline
\makecell[l]{\textbf{Greedy}\\ \textbf{Model}}& - & \makecell[c]{ \textbf{Ours} } & Greedy\\
\cline{1-3}
 \multirow{1}{*}{\makecell[l]{\textbf{Markov}\\\textbf{Model}}} & \multirow{2}{*}{\makecell[l]{\cite{DBLP:conf/emnlp/NgL04}, \\ \cite{DBLP:conf/acl-sighan/LowNG05}}}& MMTNN: \cite{pei2014max}  &  \\
\cline{3-4}
&&\makecell[l]{\cite{DBLP:conf/emnlp/ZhengCX13}, \\LSTM: \cite{chen2015long}} &\multirow{1}{*}{Viterbi}\\
\cline{1-3}
\makecell[l]{\textbf{Sequence}\\ \textbf{Labeling}\\ \textbf{Model}}&\makecell[l]{CRF: \cite{peng2004chinese}, \\ semi-CRF: \cite{DBLP:conf/emnlp/Andrew06}, \cite{DBLP:conf/naacl/SunZMTT09} }& \makecell[l]{CNN+CRF:\cite{wang2017convolutional}, \\  BiLSTM+CRF:\cite{DBLP:conf/emnlp/MaGW18} }&\\
\hline
\makecell[l]{\textbf{General}\\\textbf{ Graph}\\\textbf{ Model}}& \makecell[l]{ \cite{DBLP:conf/acl/ZhangC07} } & \makecell[l]{ LSTM+GCNN: \cite{aclCaiZ16}, \\ LSTM+GCNN: \cite{CaiZZXWH17}\\ \cite{DBLP:conf/aaai/WangCLXZS19} }&\makecell[l]{Beam\\ search}\\
\hline
\end{tabular}}
\caption{The classification of Chinese word segmentation model.}
\label{class1}
\end{table*}

Learning from an annotated corpus with segmentation, the CWS task may be generally modeled as a decoder which performs segmentation based on a scoring module in terms of contextual feature based representations.
Table \ref{class1} summarizes typical CWS models according to their decoding ways. Markov models such as \cite{DBLP:conf/emnlp/NgL04} and \cite{DBLP:conf/emnlp/ZhengCX13} depend on the maximum entropy model or maximum entropy Markov model both with Viterbi decoding. Besides, conditional random field (CRF) or Semi-CRF for sequence labeling has been used for both traditional and neural models though with different representations \cite{peng2004chinese,DBLP:conf/emnlp/Andrew06,wang2017convolutional,DBLP:conf/emnlp/MaGW18}.   

\begin{table*}[t]
	\small
	\centering
	\begin{tabular}{|c|c|c|c|}
	\hline
	\multicolumn{2}{|c|}{Models} &Characters&Words\\
	\hline
	\multirow{3}{*}{character based}&\multicolumn{1}{c|}{Ours} & $c_0,c_1,\ldots,c_i,c_{i+1},\ldots,c_{n}$ &-\\	
	\cline{2-4}
	&\cite{DBLP:conf/emnlp/ZhengCX13}, \ldots&$c_{i-2},c_{i-1},c_{i},c_{i+1},c_{i+2}$&-\\
	\cline{2-4}
	&\cite{chen2015long}&$c_0,c_1,\ldots, c_i,c_{i+1},c_{i+2}$&-\\
	\hline
	\multirow{2}{*}{word based} &\cite{DBLP:conf/acl/ZhangC07}, \ldots&$c \text{ in } w_{j-1},w_j,w_{j+1}$&$w_{j-1},w_j,w_{j+1}$\\
	\cline{2-4}
	&\cite{aclCaiZ16,CaiZZXWH17}&$c_0,c_1,\ldots ,c_{i}$&$w_0,w_1,\ldots ,w_{j}$\\
	\hline	
	\end{tabular}
	
	\caption{Feature windows of different models. $i(j)$ is the index of current character(word).}
\label{feature-window}
\end{table*}

Recent neural CWS research have been concerned about the following three perspectives \cite{Emerson2005}.

\textbf{Decoder}. As CWS is a kind of structure learning task, the decoder module generally determines which type of detailed algorithm should be adopted for segmentation, also it may limit the capability of defining feature. As shown in Table 2, not all models can support the word-level features as CWS is a task to predict word boundary. Thus recent works focus on finding more general or flexible decoder design to make model learn the representation of segmentation more effective such as \cite{aclCaiZ16,CaiZZXWH17}. 

\textbf{Encoder.} Practice in various natural language processing tasks has shown that effective representation is essential to the performance improvement. For such a module in neural models, it is more than an encoder now, which is regarded as the most improvement perspective against traditional models. Thus for better CWS, it is crucial to encode the input character, word or sentence into a distinguishable representation. Table \ref{feature-window} summarizes regular feature sets for typical CWS models including ours as well. The building blocks that encoders use include recurrent neural network (RNN) and convolutional neural network (CNN), and long short-term memory (LSTM) network.

\textbf{External resources and pre-trained embedding.}  
Using external resource such as pre-trained embeddings or language representation provides an alternative for performance improvement other than designing better models \cite{Yang2017Neural}. SIGHAN Bakeoff therefore defines two types of evaluation settings, closed test limits all the data for learning  not to be beyond the given training set, while open test does not take this limitation \cite{Emerson2005}. This work will focus on the closed test setting by finding a better model design for further CWS.

Generally speaking, both the major difference between traditional and neural models, and what mostly distinguishes the neural models are about the way to represent input sentences, while the options of decoding algorithms are bounded to how to formalize the CWS into a structural learning task. As shown in Table \ref{class1}, using Markov contextualized features, Markov models and CRF-based models are capable of using Viterbi decoders with polynomial time complexity. Furthermore, to accommodate more rich features means that the model has to take a deeper structural learning which also requires more complex decoding algorithms \cite{DBLP:conf/acl/ZhangC07,aclCaiZ16}. However, for such a case, deterministic decoding algorithms may have an intractable complexity, thus it forces the model to use an approximate beam search strategy luckily with low-order polynomial time complexity $O(Mnb^2)$, where $b$ is beam width,$n$ is the sentence size, and $M$ is a constant representing the model complexity. When the beam width $b$=1, the beam search will reduce to greedy algorithm with a much better time complexity $O(Mn)$. 

To make the decoding practical, the beam width $b$ has to be carefully tuned for a tradeoff between accuracy and efficiency: A larger $b$ will make the learning and segmentation extremely slow, while a small $b$ cannot sufficiently guarantee the segmentation performance. However, there has long been a unheeded observation that good enough representations can offer good enough segmentation even though only using a greedy segmentation algorithm. 
\cite{Sproat-2003} create a topline evaluation by using only using vocabulary from test set to perform a greedy segmentation (maximum matching), which yields around 99\% F-scores on all datasets. For neural models, \cite{CaiZZXWH17}  verify that if the representations are good enough, beam width 1 can still give state-of-the-art performance compared to their early model with a full beam search decoder in \cite{aclCaiZ16}. Therefore, undertaking a fixed greedy segmentation algorithm, this paper only focuses on more effective encoder design for even better representation.
 
Our model only consists of attention mechanisms as building blocks plus two highway connections via a virtual hidden layer for smooth training. 
Our model is simply stacked by a variant of Transformer encoder \cite{VaswaniSPUJGKP17} and a biaffine attention scorer \cite{dozat2017deep}. Empowered by the self-attention mechanism, the Transformer has been good at capturing long-range dependencies for input sentence. We propose Gaussian-masked Directional (GD) multi-head attention to facilitate the learning of localness, position and directional information for CWS, so that we have the proposed GD-Transformer.

With our further improved encoder, our model uses only simple unigram features to generate representation of sentences for scoring.
Our model will be strictly evaluated on benchmark datasets from SIGHAN Bakeoff shared task  in terms of closed test setting, and experimental results show that our model achieves new state-of-the-art. 

The technical contributions of this paper can be  summarized as follows.

$\bullet$ To especially enhance the representation of localness information and directional information, we propose a new Gaussian-masked Directional Transformer encoder.

$\bullet$ Motivated from a simple design idea, we present a new CWS model which is stacked with only attention blocks. 

$\bullet$ With a powerful enough encoder, for the first time, we show that unigram (character) features plus greedy segmentation algorithm can support yielding strong performance instead of using diverse $n$-gram (character and word) features and highly complex decoding algorithms.

\section{Related Work}

\cite{xue2003chinese} first formalize CWS as a sequence labeling task, considering CWS as a supervised learning from annotated corpus with human segmentation. \cite{peng2004chinese} further adopt standard sequence labeling tool CRFs for CWS modeling, achieving new state-of-the-art. 
\cite{zhao2006effective} show that different character tag sets can make essential impact for segmentation performance. \cite{zhao-etal-2006-improved} propose a CWS system developed for Bakeoff-2006 based on CRF, which is based on their proposed 6-tag set for character position tagging and achieved state-of-the-art performance at then. \cite{Zhao07incorporatingglobal} present a novel Character tagging based CRF framework which is capable of exploiting global information for performance enhancement. 

Neural word segmentation has been widely used to minimize the efforts in feature engineering. \cite{DBLP:conf/emnlp/ZhengCX13} first introduce the neural model into CWS with sliding-window based sequence labeling.  \cite{chen2015long} use LSTM to enhance the learning of long distance information. 

However, introducing neural models themselves does not really introduce substantial performance improvement in terms of strict closed test of SIGHAN Bakeoff according to \cite{zhao2017}. Most researchers actually seek help from joint learning, extra learning resources including dictionaries, pre-trained embedding, deeper information extracted from training set and so on.
(1) For joint learning,
\cite{DBLP:conf/aaai/LyuZJ16} explore a joint model that performs segmentation, POS-Tagging and chunking simultaneously.
 \cite{DBLP:conf/ijcai/ZhangFY17} present a joint model to enhance the segmentation of Chinese microtext by performing CWS and informal word detection simultaneously.
(2) For extra resources or clues, 
\cite{DBLP:conf/aaai/WangCLXZS192} propose to incorporate unlabeled and partially-labeled data.    

Only a few researches are known for concentrating on strengthening the model itself. To accommodate more rich features through a more broadly structural modeling
\cite{aclCaiZ16} propose a neural framework that eliminates context windows and utilize complete segmentation history. 
\cite{wang2017convolutional} propose a character-based convolutional neural model to capture $n$-gram features automatically and an effective approach to incorporate word embeddings. \cite{CaiZZXWH17} further improve the model in \cite{aclCaiZ16} and show that a greedy  segmenter can perform fast and accurately in terms of only presenting effective representations. This work follows this line of research by offering even strengthened model design from simple idea, including the least building block type for encoder (attention only), the least feature type for scoring (unigram only) and the least computational complexity for decoding (greedy segmentation).

The original Transformer encoder consists of a stack of \textbf{N} identical layers and each layer has one multi-head self-attention layer and one position-wise fully connected feed-forward layer \cite{VaswaniSPUJGKP17}. One residual connection is around two sub-layers and followed by layer normalization. Several variants are proposed to enhance ability of capturing the localness relationship. \cite{shaw-etal-2018-self} propose an effcient way to incorporate relative and absolute position representation. \cite{yang-etal-2018-modeling} cast localness modeling as a learnable Gaussian bias to enhance the ability of capturing useful local context. \cite{DBLP:conf/icassp/KimEL20} propose a Transformer with Gaussian-weighted self-attention to improved speech-enhancement performance. \cite{DBLP:conf/aaai/0001WZDZ020} propose using syntax to guide the text modeling based on self-attention network sponsored Transformer-based encoder. Transformer based pre-trained language models have become a standard performance enhancement means for various NLP tasks \cite{DBLP:conf/aaai/0001WZLZZZ20}.

\section{Models}

\begin{figure}
\hspace*{-1cm}
\includegraphics[scale=0.18]{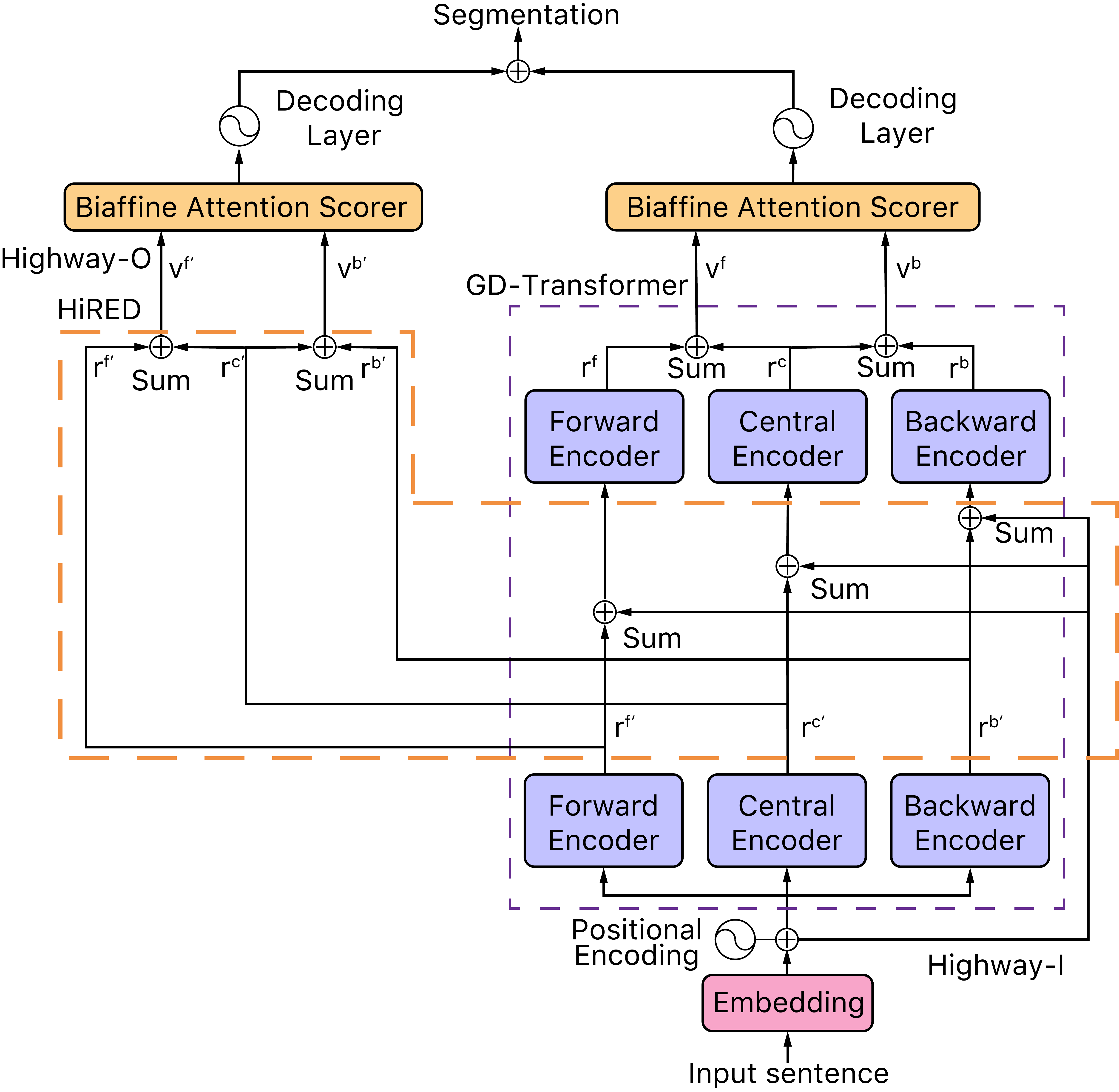}
\caption{The architecture of our model.}
\label{mods}
\end{figure}

Our model for CWS task is composed of an encoder to represent the input and a decoder based on the encoder to perform actual segmentation. Figure \ref{mods} is the architecture of our model. The model feeds sentence into encoder. Embedding captures the vector $e$ of the input character sequences of $c$. The encoder maps vector sequences of $ \boldmath{e}$ to two sequences of vector which are $ \boldmath{v^b}$ and $\boldmath{v^f}$ as the representation of sentences. With $v^b$ and $v^f$, the biaffine scorer scores each segmentation gaps which makes our decoder is as simple as one layer, using a threshold to directly and greedily predict every word boundaries of the input. 

\subsection{Gaussian-Masked Directional Transformer}

The standard Transformer encoder consists of a stack of \textbf{N} identical layers and each has one multi-head self-attention layer and one position-wise fully connected feed-forward layer. One residual connection is around two sub-layers and followed by layer normalization \cite{VaswaniSPUJGKP17}. 

The proposed Gaussian-masked Directional (GD) Transformer encoder adopts two key architecture revisions over the standard  Transformer. (1)
Our encoder includes three parallel directional encoding pipelines instead of only one bidirectional encoder in the original Transformer. 
(2)
By replacing the standard multi-head self-attention with the proposed Gaussian-masked Directional (GD) multi-head self-attention which captures representations from different directions, the resulted encoder may gain better ability of capturing the localness information and position information for the importance of adjacent characters.

\paragraph{Encoder Stacks}
In CWS task, word boundary forms a gap between two adjacent characters and divides one sequence into two parts, one part in front of the gap and one part in the rear of it. The forward encoder and backward encoder are proposed to capture information of two directions which correspond to two parts divided by the gap. Assuming that one unidirectional encoder can capture information from one particular direction, we stack three parallel encoding modules, forward, backward and center encoders as shown in Figure \ref{mods}.  

The central encoder is to capture information from both directions, which is with the same architecture as the original Transformer. 
Standard scaled dot-product attention matrix is calculated by dotting query $Q$ with all keys $K$. For the forward encoder, we forcibly set all values inside the attention matrix representing the character pair relation after the concerned character as 0 so that the encoder can focus on the forward characters. For the backward encoder, we take the similar matrix value setting operations.

The encoder respectively outputs one forward and one backward representations for each position, and then both are fused with the representation given by the center encoder to form the updated forward and backward representations, respectively.

$v^{b}$ = $r^{b}$ + $r^{c}$, $v^{f}$ = $r^{f}$ + $r^{c}$,

\noindent where $v^{b}$ and $v^{f}$ represent the backward and forward representation, respectively, $r^{b}$, $r^{c}$ and $r^{f}$ are representations from backward encoder, center encoder and forward encoder, respectively.

\paragraph{Gaussian-Masked Directional Multi-Head Attention}
Similar as scaled dot-product attention in the original Transformer \cite{VaswaniSPUJGKP17}, our proposed Gaussian-masked directional attention can be described as a function to map queries and key-value pairs to the representation of input. Here queries, keys and values are all vectors. Standard scaled dot-product attention is calculated by dotting query $Q$ with all keys $K$, dividing each values by $\sqrt{d_k}$, where $\sqrt{d_k}$ is the dimension of keys, and apply a softmax function to generate the weights in the attention:
\begin{equation}
\label{orisof}
Attention(Q, K, V) = softmax(   \frac{Q{K^T}}{\sqrt{d_k}})V
\end{equation}

Different from scaled dot-product attention, Gaussian-masked directional attention expects to pay attention to the adjacent characters of each positions and cast the localness relationship between characters as a fix Gaussian weight for attention. We assume that the Gaussian weight only relies on the distance between characters.

Firstly we introduce the Gaussian weight matrix $G$=($g_{ij}$) which presents the localness relationship between each two characters:

\begin{equation}
\label{gausweis2}
\\g_{ij}= \Phi(dis_{ij}) = \sqrt{\frac{2}{{\sigma}^{2}\pi}}\int_{- \infty}^{-dis_{ij}}{exp(-\frac{x^2}{2{\sigma}^2})dx}
\end{equation}
where $g_{ij}$ is the Gaussian weight between character $i$ and $j$, $dis_{ij}$ is the distance between character $i$ and $j$, $\Phi(x)$ is the cumulative distribution function of Gaussian, $\sigma$ is the standard deviation of Gaussian function and it is a hyperparameter in our method. Eq. (\ref{gausweis2}) ensures the Gaussian weight equals 1 when $dis_{ij}$ is 0. The larger distance between characteristics, the smaller the weight is, which lets one character affect its neighbors more than those non-neighbors.

To combine the Gaussian weight to the self-attention, we produce the Hadamard product of Gaussian weight matrix $G$ and the score matrix produced by $Q{K^{T}}$
\begin{equation}
\begin{aligned}
AG(Q, K, V)=softmax(\frac{Q{K^T} * G}{\sqrt{d_k}})V
\end{aligned}
\label{wgausweis}
\end{equation}
where $AG$ as the Gaussian-masked attention ensures that 
adjacent characters have a stronger relationship than those non-neighbored ones.

The scaled dot-product attention models the relationship between two characters without regard to their distances in one sequence. For CWS task, the weight between adjacent characters should be more important while it is hard for self-attention to achieve the effect explicitly because the self-attention cannot get the order of sentences directly. The Gaussian-masked attention adjusts the weight between characters and their adjacent character to a larger value which stands for the effect of adjacent characters.

\begin{figure}[!htbp]
\centering
\subfigure[The architecture of Gaussian-masked directional multi-head attention.]{
\begin{minipage}[t]{0.45\linewidth}
\centering
\includegraphics[scale=0.25]{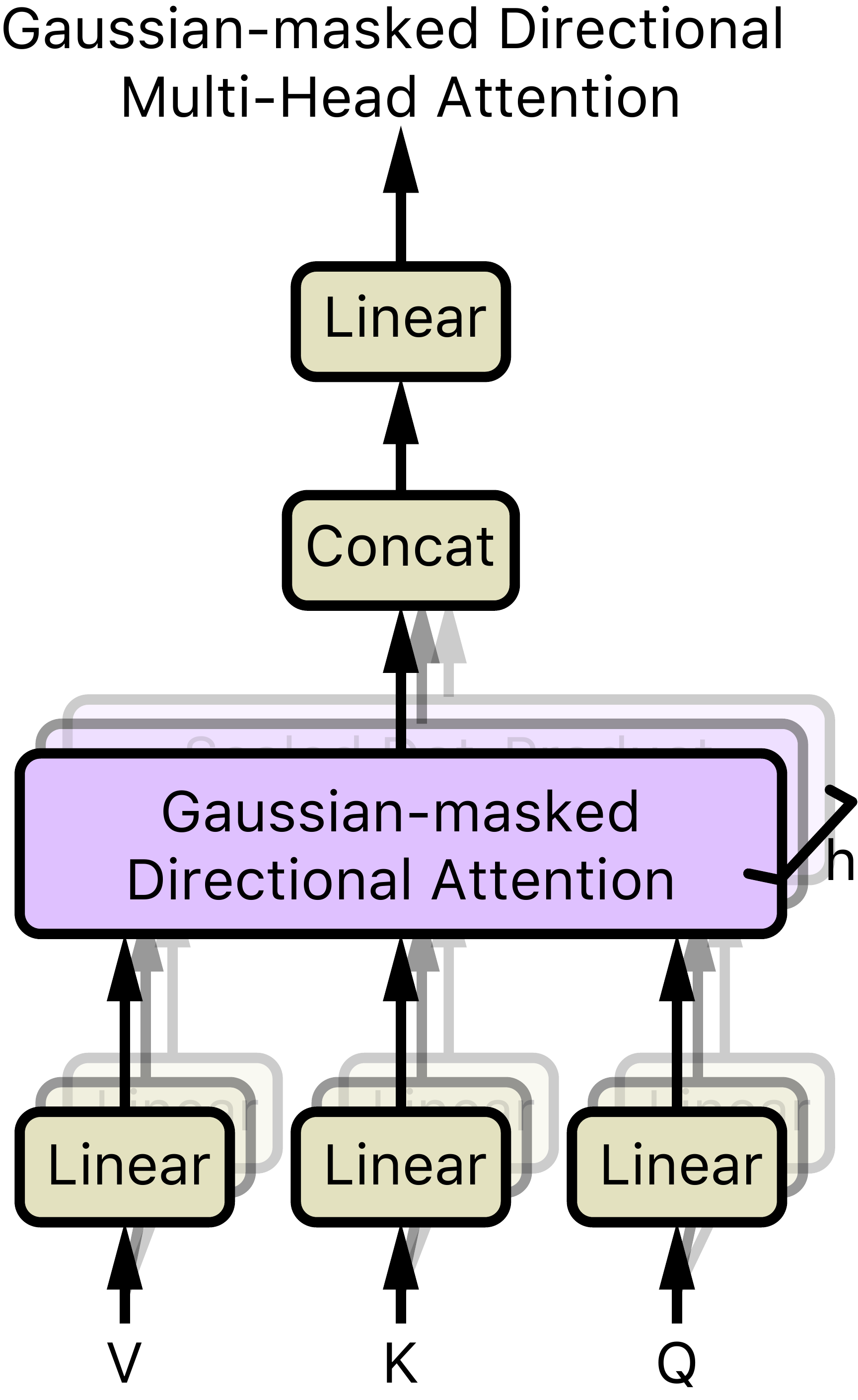}

\label{encoders}
\end{minipage}
}
\hfill
\subfigure[The Gaussian-masked directional attention.]{
\begin{minipage}[t]{0.45\linewidth}
\centering
\includegraphics[scale=0.17]{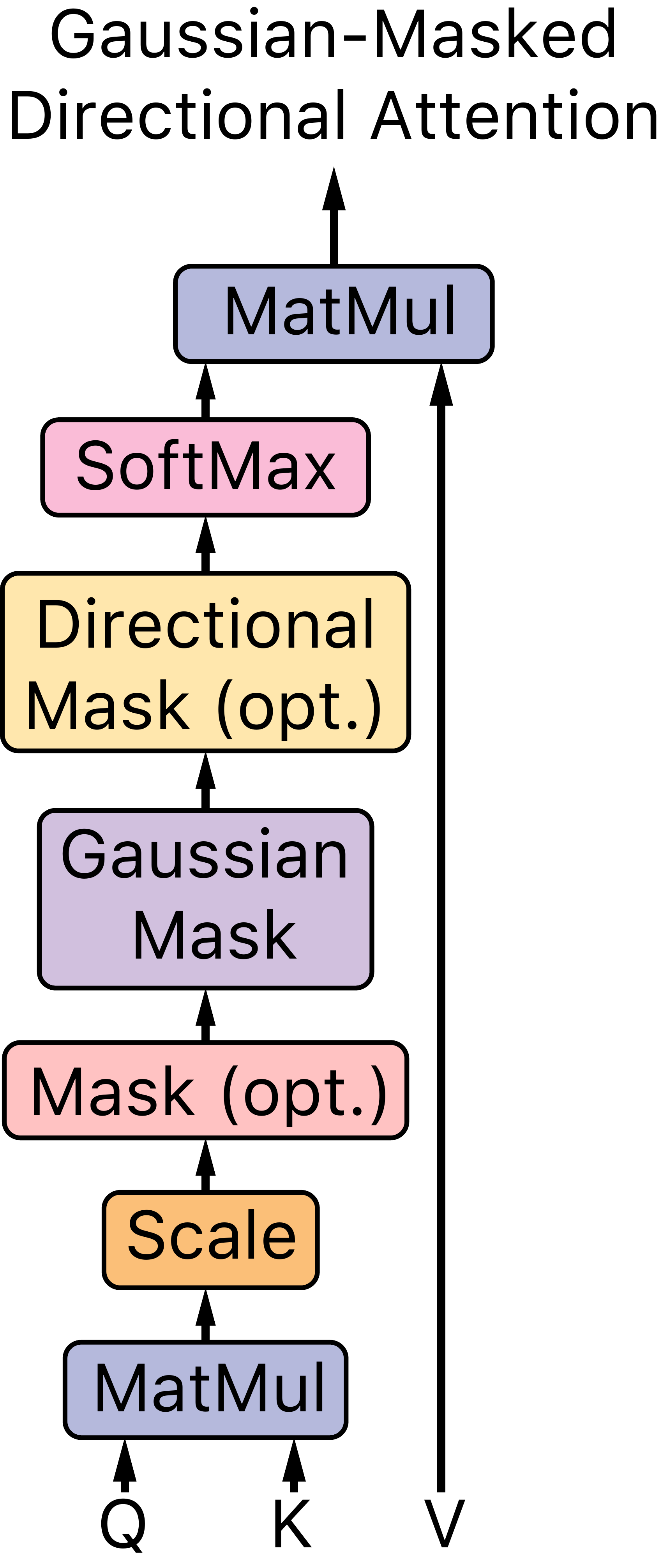}
\label{gausae2}
\end{minipage}
}
\caption{Illustration of Gaussian-masked directional multi-head attention.}
\end{figure} 

For forward and backward encoder, the self-attention sub-layer needs to use a triangular matrix mask to let the self-attention focus on different weights:
\begin{equation}
\label{forward}
\begin{aligned}
g^{f}_{ij} = \left\{
\begin{aligned}
g_{ij} & , & pos_j \le pos_i, \\
-\infty & , & others.
\end{aligned}
\right.\\
g^{b}_{ij} = \left\{
\begin{aligned}
g_{ij} & , & pos_i \le pos_j, \\
-\infty & , & others.
\end{aligned}
\right.
\end{aligned}
\end{equation}
where $pos_i$ is the position of character $c_i$. The triangular matrix for forward and backward encode are:
$\left[ \begin{matrix}
   1 &  0 &  \cdots &0\\
   1 & 1 &  \cdots &0\\

\vdots&\vdots &\ddots&\vdots\\
 1 & 1 &  \cdots & 1\\
  \end{matrix} \right]$ $\left[ \begin{matrix}
   1 &  1 &  \cdots &1 \\
   0 & 1 &  \cdots &1 \\

\vdots&\vdots &\ddots&\vdots\\
 0 & 0 &  \cdots & 1\\
  \end{matrix}\right]$

Similar as \cite{VaswaniSPUJGKP17}, we use multi-head attention to capture information from different dimension positions as Figure \ref{encoders} and get Gaussian-masked directional multi-head attention $GMH$ as follows, 
\begin{equation}
\begin{aligned}
GMH(Q,K,V)&=Concat(head_1,...,head_h)W_m,\\
head_i&=AG(QW_i^q,KW_i^k,VW_i^v)
\end{aligned}
\label{mulsa}
\end{equation} 
where 
${W_i^q, W_i^k,W_i^v} \in \mathbb{R}^{d_k \times d_h}$ is the parameter matrices to generate heads, $W_m$ is a parameter matrices of $\mathbb{R}^{d_k \times d_k}$ to generate the attention, $d_k$ and $d_h$ are dimensions of model and one head, respectively.

\subsection{Biaffine Attention Scorer}
\label{biaffinal-scorer-secs}
Our model straightforwardly predicts gap between two adjacent characters as word boundary or not. In detail, we set a label value 1 to indicate word boundary, and 0 means no word boundary.
Such a gap labeling task thus requires information of the two adjacent characters. In the meantime, the relationship between adjacent characters can be represented as the gap label. 

Biaffine attention scorer is used to label the gap \cite{dozat2017deep, li-etal-2018-joint-learning,cai-etal-2018-full,zhou-zhao-2019-head,he-etal-2019-syntax}. 
The distribution of labels in a labeling task is often uneven. Biaffine attention uses bias terms to alleviate the burden of the fixed bias term and get the prior probability which makes it different from bilinear attention. The distribution of the gap is uneven that is similar as other labeling task, which makes biaffine available for our task. 

Biaffine attention scorer labels the target depending on information of  independent unit and the joint information of two units. In biaffine attention, the score $s_{ij}$ of characters $c_i$ and $c_j$ $(i < j)$ is calculated by:
\begin{equation}
\label{biaffines}
\begin{aligned}
s_{ij} &= BiaffinalScorer(v_i^f,v_j^b) \\
&=(v_i^f)^TWv_j^b + U(v_i^f \oplus v_j^b) + b\\
\end{aligned}
\end{equation}
where $v_i^f$ and $v_i^b$ represent respectively the forward and backward information of $c_j$, $W$, $U$ and $b$ are all learnable parameters. $W$ is a matrix with shape $(d_i \times N\times d_j)$ and $U$ is a $(N\times (d_i + d_j))$ matrix where $d_i$  is the dimension of vector $v_i^f$ and $N$ is the number of labels.
\begin{figure}[!htb]
\centering
\small
\includegraphics[scale=0.17]{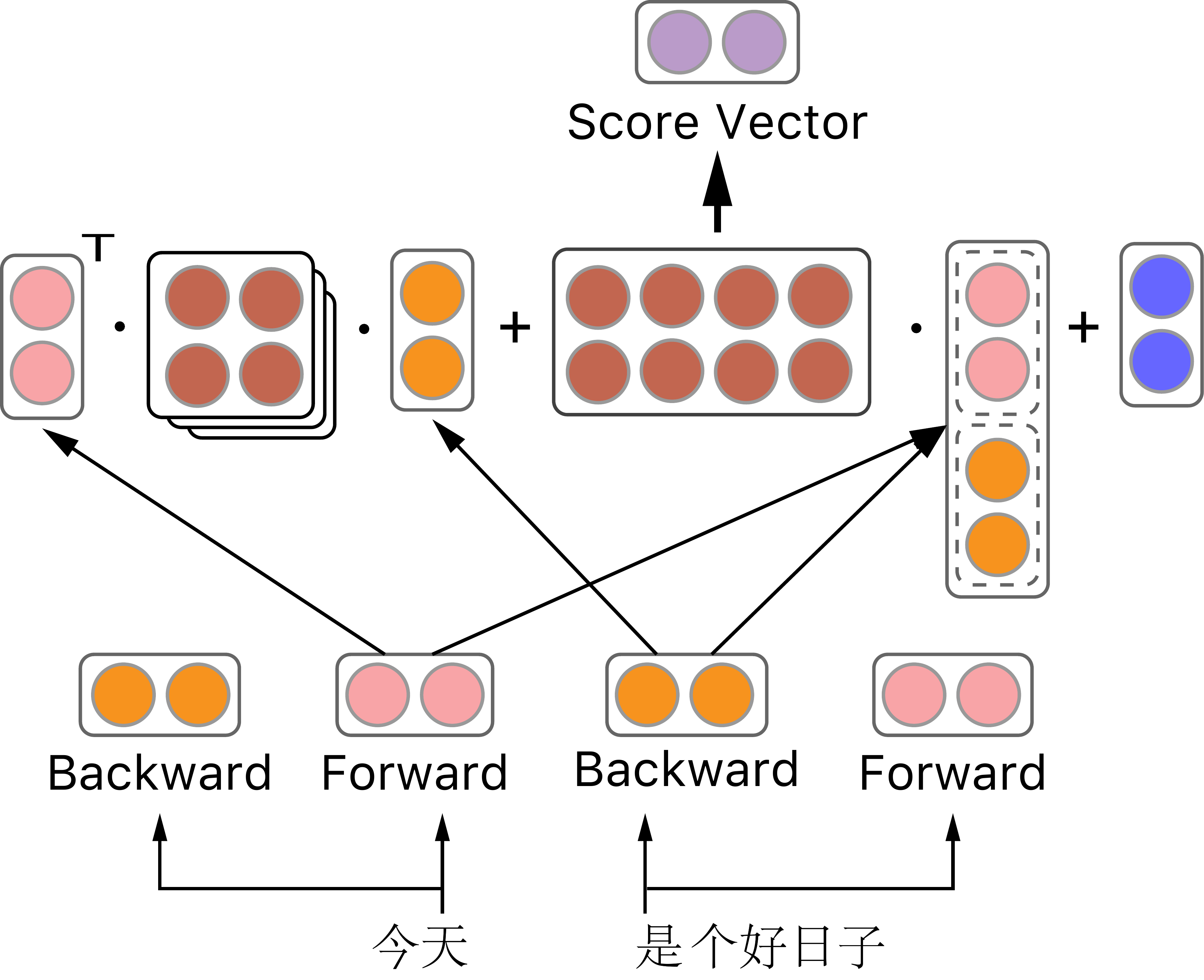}
\caption{An example of biaffine scorer labeling the gap. The biaffine attention scorer only uses the forward information of front character and the backward information of character to label the gap.}
\label{biaffinescore}
\centering
\end{figure}

In our model, the biaffine scorer uses both the forward and backward character information on either side of the gap to distinguish the position of characters. Figure \ref{biaffinescore} is an example of gap labeling. The bidirectional scoring ensures that the boundaries of words can be determined by adjacent characters with different directional information. The score vector of the gap is formed by the probability of being a boundary of word. Further, the model generates all boundaries using activation function in a greedy decoding way.   

\subsection{Highway Connections via Hidden Layer}

To smooth the training and fully exploit representations from hidden states, we additionally introduce
two Highway connections \cite{srivastava2015highway} via a virtual hidden layer which is called Hidden Representations for Early Decoding (HiRED) in the middle of the Transformer encoder. In our model design, we always put the HiRED layer in the central position among all layers of the encoder, thus the HiRED layer divides each directional encoder (forward, backward or center) pipelines into two parts (front and rear) as shown in Figure \ref{mods}.

For the highway connection specifications,
the first connection (called Highway-I) respectively feeds the input embedding to the rear pipelines of the three directional encoders by adding into the embeddings from HiRED layer. Suppose that three front directional encoders respectively give encoding output, ${r^f}'$, ${r^c}'$ and ${r^b}'$. Then the corresponding three rear directional encoders will receive input as  $e+{r^f}'$, $e+{r^c}'$ and $e+{r^b}'$. 
To feed
the second connection (called Highway-O), we perform the same summing as the main encoder output,

${v^{b}}'$ = ${r^{b}}'$ + ${r^{c}}'$, ${v^{f}}'$ = ${r^{f}}'$ + ${r^{c}}'$,

\noindent then let ${v^f}'$ and ${v^b}'$ as the HiRED output go through another same biaffine scorer and a decoder as that of the main encoder. The two decoder layers together give a sum loss for the entire model. 

Biaffine attentin scorer makes it possible to generate a segmentation by using output of HiRED with little cost during training. With this segmentation, we add representation of characters which belong to the same word together and get a new vector, which plays a similar role as a word embedding. This vector will be fed to encoder layer behind HiRED directly. The operations in HiRED layer can also be viewed as one attention. It makes the model focus on adjacent characters which may be likely in one word.

\subsection{Training Objective}
The training target of our model is to let the biaffine attention scorer approach the the gold score vector according to the gold segmentation. We adopt cross entropy (CE) loss for training,
\begin{equation}
\begin{aligned}
q_i^j&=-s_{i,i+1}^j+{\rm log}({\rm exp}(s_{i,i+1}^0)+{\rm exp}(s_{i,i+1}^1)),\\
CE&=\frac{1}{l}\sum_{i=1}^{l}(q_i^1p+q_i^0(1-p))\nonumber
\end{aligned}
\label{celoss}
\end{equation}
where $q^j_i$ is the log-probability of the $i$-th gap labeled as $j$$\in$\{1,0\}. Here 1 indicates word boundary and 0 means not. $s_{i,i+1}^j$ is the biaffine score of $i$-th gap labeled as $j$. $p$ is the ground-truth probability which can only be 0 or 1. $l$ is the number of gaps in one input sentence.

\begin{table}[!htb]
\centering
\small
\scalebox{0.9}{
\begin{tabular}{|l|r|r|r|r|}
\hline
&\multicolumn{1}{c|}{\textbf{PKU}} & \multicolumn{1}{c|}{\textbf{MSR}} \\
\hline
\textbf{Sentences}&19,056&86,924\\
\textbf{Max length (Character)}&1019&581 \\
\textbf{Max length (Word)}&659&338 \\
\textbf{Word  Types} &55,303 & 88,119 \\
\textbf{Words} &1,109,947 & 2,368,391 \\
\textbf{Character Types} &4,698&5,167\\
\textbf{Characters}  & 1,826,448 &4,050,469 \\
\hline

& \multicolumn{1}{c|}{\textbf{AS}}& \multicolumn{1}{c|}{\textbf{CITYU}} \\

\hline
\textbf{Sentences}&708,953&53,019\\
\textbf{Max length (Character)}&188&350\\
\textbf{Max length (Word)}&211&85\\
\textbf{Word  Types} & 141,340 & 69,085 \\
\textbf{Words} & 5,449,698 &1,455,629\\
\textbf{Character Types} &6,117&4,923\\
\textbf{Characters}  &8,368,050&2,403,355\\
\hline
\end{tabular}
}
\caption{Statistics of SIGHAN Bakeoff 2005 datasets.}
\label{datas2}
\end{table}

\begin{table}[!htb]
\small
\centering
\begin{tabular}{|l|r|}
\hline
&\multicolumn{1}{c|}{\textbf{Parameters}} \\
\hline
\textbf{dimension of hidden vector} &256   \\
\textbf{number of layer} &6  \\
\textbf{dimension of FF} &1024\\
\textbf{dropout}  & 0.1\\
\textbf{warmup} & 8000 \\
\textbf{number of head}&4\\
\textbf{batch size}&4096\\
\hline
\end{tabular}
\caption{Hyperparameters.}
\label{hyper-parameters}
\end{table}

\begin{table*}[t]
			\renewcommand\tabcolsep{5.0pt}
			\centering
			\small
			\begin{tabular}{l|ccc|ccc|ccc|ccc}
				\hline
				\multirow{3}{*}{Models}& \multicolumn{3}{c|}{PKU} &\multicolumn{3}{c|}{MSR}& \multicolumn{3}{c|}{AS} &\multicolumn{3}{c}{CITYU}\\
				\cline{2-13}
				&  \makecell[c]{$\text{F}_1$\\ } &\makecell[c]{Tr.\\(hours)}&\makecell[c]{Test \\(sec.) }&\makecell[c]{$\text{F}_1$\\ } &\makecell[c]{Tr.\\(hours)}&\makecell[c]{Test \\(sec.) }&  \makecell[c]{$\text{F}_1$\\ } &\makecell[c]{Tr.\\(hours)}&\makecell[c]{Test \\(sec.) }&\makecell[c]{$\text{F}_1$\\ } &\makecell[c]{Tr.\\(hours)}&\makecell[c]{Test \\(sec.) }\\
				\hline
				\cite{chen2015long} & \textbf{95.7}&58&105&96.4&117 &120&-&-&-&-&-&- \\
				\cite{aclCaiZ16} &95.2&48&95&96.4&96&105&-&-&-&-&-&-\\
				\cite{CaiZZXWH17}&95.4&\textbf{3}&25&97.0 &  \textbf{6}&30&95.2&-&-&95.4 &  - &-\\
				\cite{DBLP:conf/emnlp/ZhouYZHDC17}&95.0&-&-&97.2 &-&-&-&-&-&-&-&-\\
				  \cite{DBLP:conf/emnlp/MaGW18}&95.4&-&-&97.5&-&-&95.5&-&-&95.7&-&-\\
				\cite{DBLP:conf/aaai/WangCLXZS19}&\textbf{95.7}&-&-&97.4&-&-&95.6&-&-&\textbf{95.9}&-&-\\
				\hline
				Our results &95.5&33&\textbf{4}&\textbf{97.6}&15&\textbf{4}&\textbf{95.7}&\textbf{67}&\textbf{10}&95.4&\textbf{17}&\textbf{1.5}\\
				\hline
			\end{tabular}
\caption{Results on SIGHAN Bakeoff datasets in closed test. - indicates there is no reported result in the corresponding paper. (Tr.: Training). } 
\label{pkums}
\end{table*}

\section{Experiments}
\subsection{Experimental Settings}

\paragraph{Data}
Our models are trained and evaluated on benchmark datasets from SIGHAN Bakeoff 2005 \cite{Emerson2005} which has four datasets, PKU, MSR, AS and CITYU. Table \ref{datas2} shows the statistics of train data. F-score is to evaluate the performance.

\paragraph{Embedding Initialization} 
Our model only adopts unigram features, so we only train character embeddings. On closed test, we use embeddings initialized randomly. On open test, our character embeddings are pre-trained on Chinese Wikipedia corpus by word2vec \cite{Mikolov} toolkit. The corpus for pre-training is converted to simplified Chinese\footnote{OpenCC is used to transfer data from traditional Chinese to simplified Chinese, available at https://github.com/BYVoid/OpenCC.} and trivially segmented into characters. 
\paragraph{Hyperparameters}

Our hyperparameter settings are  in Table \ref{hyper-parameters}. 
All the settings are
tuned on development sets\footnote{Following conventions, the last 10\% sentences of training
corpus are used as development set.}. We set the standard deviation of Gaussian function in Eq. (\ref{gausweis2}) to 2. Each training batch contains sentences with at most 4096 tokens.

\paragraph{Optimizer}
To train our model, we use the Adam \cite{DBLP:journals/corr/KingmaB14} optimizer with $\beta_1=0.9$, $\beta_2=0.98$ and $\epsilon=10^{-9}$. The learning rate schedule is the same as \cite{VaswaniSPUJGKP17}:

$lr$ = $d^{-0.5}$ $\cdot$ min($step^{-0.5}$, $step$ $\cdot$ $warmup_{step}^{-1.5}$)

\noindent where $d$ is the dimension of embeddings, $step$ is the step number of training and $warmup_{step}$ is the step number of warmup. When the number of step is smaller than the step of warmup, the learning rate increases linearly and then decreases. 

\paragraph{Hardware and Implements}
Our models are trained on a single CPU (Intel i7-5960X) and an nVidia 1080 Ti GPU, in terms of an implementation using Pytorch 1.0\footnote{Code is available at: \url{https://github.com/akibcmi/SAMS}}.

\subsection{Results}

Tables \ref{pkums} compares recent models and ours in terms of closed test setting, showing that our model achieves new state-of-the-art and outperforms all the other models in MSR and AS. In the meantime, our model can achieve state-of-the-art efficiency.

Our models are also compared to the latest neural models in terms of open test setting in which any external resources, especially pre-trained embeddings or language models are allowedly used. Table \ref{datas5} shows that our models get comparable results in AS and MSR though unremarkable ones in CITYU and PKU.

However, it is well known that comparing models accurately is hard for open test setting. Though external strengths like pre-trained embeddings or models can indeed improve the performance, it is difficult to determine which factor exactly makes such a contribution, the model itself, the resource or the better using of the resource. In terms of closed test setting, that is also the reason why this work keeps focusing on improvement of the model design itself.

\begin{table}[!htb]
\centering

\setlength{\tabcolsep}{0.9mm}{
\scalebox{0.9}{
\begin{tabular}{l|c|c|c|c}
\hline
&\multicolumn{1}{c|}{\textbf{PKU}} & \multicolumn{1}{c|}{\textbf{MSR}} & \multicolumn{1}{c|}{\textbf{AS}}& \multicolumn{1}{c}{\textbf{CITYU}} \\
\hline

\cite{CaiZZXWH17} &95.8 & 97.1 & 95.3 &95.6\\
\cite{ChenSQH17}&94.3&96.0&94.6&95.6\\
\cite{wang2017convolutional}&95.7&97.3&-&-\\
\cite{DBLP:conf/emnlp/ZhouYZHDC17}&96.0&97.8&-&-\\

\cite{DBLP:conf/emnlp/MaGW18}&96.1&\textbf{98.1}&96.2&97.2\\
\cite{DBLP:conf/aaai/WangCLXZS19}&96.1&97.5&-&-\\
\cite{DBLP:journals/corr/abs-1903-04190}& \textbf{96.6}&97.9&\textbf{96.6}&\textbf{97.6}\\ 

\hline
\textbf{Our Method}&95.5&97.7&95.7&96.4\\
\hline
\end{tabular}}
}

\caption{F1 scores in open test.}
\label{datas5}
\end{table}

Compared with other LSTM models, our model performs better in AS and MSR than in CITYU and PKU. We attribute the performance difference to the impact of dataset sizes. Namely, the larger size is, the better model performs. For small corpus, the model tends to be overfitting.

Table \ref{pkums} also shows the decoding time in different datasets. Our model finishes the segmentation with the least decoding time in all four datasets, thanks to the architecture of model which only takes attention mechanism as basic block, only adopts unigram features and a greedy decoding strategy from the very beginning.

\subsection{Ablation Studies}
This subsection presents ablation studies on MSR and PKU datasets to verify the benefits of each individual component in our model\footnote{Following \cite{CaiZZXWH17}, we show the results on the respective test set for either dataset, as SIGHAN Bakeoff did not provide official development sets.}.

\paragraph{Gaussian-masked Directional Transformer.}  Table \ref{datasselfattention} gives the result of model with different Gaussian-masked directional self-attention. The third column and the fifth column are the difference of performance between GD-Transformer and other models. The results show that our full model GD-Transformer significantly outperforms the original Transformer by a large performance margin. Removing either Gaussian mask or directional mask will put negative impact over the performance of our model, which shows that both masks are indispensably necessary for our model performance.

\begin{table}[!htb]
\centering

\setlength{\tabcolsep}{0.9mm}{
\scalebox{0.9}{
\begin{tabular}{lcrcr}
\hline
&\multicolumn{2}{c}{\textbf{PKU}} & \multicolumn{2}{c}{\textbf{MSR}}\\
\hline
GD-Transformer& 95.4 &&97.6& \\

-Gaussian mask & 94.6&-0.8 &97.1&-0.5\\
-Directional mask &95.1&-0.3&97.4&-0.2\\
\hline
Transformer&94.1&-1.3&96.5&-1.1\\
\hline
\end{tabular}}
}
\caption{F1 scores on models removing different components from GD-Transformer.}
\label{datasselfattention}
\end{table}

\paragraph{Highway Connections.} Table \ref{datasdoublejointlayer} gives the results of our model respectively removing the highway connections and the related HiRED layer part, which shows that each highway takes its contribution to the overall performance. However,  the comparison shows that introducing all the components makes our model training much faster.

\begin{table}[t]
\centering

\setlength{\tabcolsep}{0.9mm}{
\scalebox{0.9}{
			\begin{tabular}{lcccccc}
				\hline
				\multirow{1}{*}{Models}& \multicolumn{2}{c}{PKU} &\multicolumn{2}{c}{MSR}\\
				\cline{2-5}
				&  \makecell[c]{$\text{F}_1$\\ } &\makecell[c]{Training\\(hours)}&\makecell[c]{$\text{F}_1$\\ } &\makecell[c]{Training\\(hours)}\\
				\hline
				Our full model& 95.5 &33&97.6&15 \\
\hline
-Highway-I &95.2&60&97.5&96\\
-Highway-O &95.3&45&97.4&102\\
\hdashline
-both highways&95.1&80&97.5&105\\
				\hline
			\end{tabular}
}}
			\caption{F1 scores and training time on models related to highway connections and HiRED layer.}
\label{datasdoublejointlayer}
\end{table}

\begin{table}[!htb]
\centering

\setlength{\tabcolsep}{0.9mm}{
\scalebox{0.9}{
\begin{tabular}{lcccc}
\hline
&\multicolumn{2}{c}{\textbf{PKU}} & \multicolumn{2}{c}{\textbf{MSR}}\\
\hline
Our full model & 95.5 &&97.6& \\
\hline
-Forward encoder & 95.3&-0.2 & 97.4&-0.1 \\
-Center encoder & 95.3 & -0.2 & 97.5& -0.1\\
-Backward encoder & 95.4&-0.1 & 97.5 &-0.2\\
\hline
\end{tabular}}
}
\caption{F1 scores of results on model removing different encoder from model.}
\label{datasbackward}
\end{table}

\paragraph{Directional Encoder.} Table \ref{datasbackward} gives the results of our models respectively removing the forward, center and backward encoders, which impacts performance of our model and shows that directional encoder and undirectional encoders are all indispensable for our model. The third column and the fifth column are the difference of performance between our full model and our models removing one encoder.

\section{Conclusion}
For Chinese word segmentation, upholding the belief that a better representation is all we need and thus taking a greedy decoder for fast segmentation as the basis, we only focus on the encoder design and propose an attention mechanism only based CWS model. Our model uses the  proposed GD-Transformer encoder to take sequence input and biaffine attention scorer to directly predict the word boundaries. To improve the ability of capturing the localness and directional information, Gaussian-masked directional multi-head attention in the GD-Transformer replaces the standard self-attention in the original Transformer. With powerful enough encoding ability,
our model only needs unigram features for scoring instead of various $n$-gram features in previous work. Our model is evaluated on standard benchmark SIGHAN Bakeoff datasets, which shows not only our model performs segmentation faster than any previous models but also gives new higher or comparable segmentation performance against previous state-of-the-art models.

\bibliographystyle{acl_natbib}
\bibliography{emnlp2020}

\end{document}